\documentclass{article} % For LaTeX2e
\usepackage{iclr2015,times}
\usepackage{hyperref}
\usepackage{url}
\usepackage{amsmath}
\usepackage{amsthm}
\usepackage{amssymb}
\usepackage{graphicx}
\usepackage{xspace}
\usepackage{tabularx}
\usepackage{arydshln}
\usepackage{wrapfig}
\usepackage[utf8]{inputenc}

\usepackage[normalem]{ulem}
\newcounter{quotecount}
\newcommand{\MyQuote}[1]{\vspace{0.4cm}\addtocounter{quotecount}{1}%
     (\arabic{quotecount})\hspace*{1cm}\parbox{12cm}{\em #1}\\[0.4cm]}

\title{Embedding Word Similarity with \\ Neural Machine Translation}

\author{
Felix Hill \\
University of Cambridge \\
\texttt{felix.hill@cl.cam.ac.uk} 
\And
KyungHyun Cho \\
Universit\'{e} de Montr\'{e}al
\And
Sébastien Jean \\
Universit\'{e} de Montr\'{e}al
\And
Coline Devin \\
Harvey Mudd College
\And
Yoshua Bengio \\
Universit\'{e} de Montr\'{e}al, CIFAR Senior Fellow
}

\iclrfinalcopy % Uncomment for camera-ready version

\begin{document}

\maketitle

\begin{abstract}
Neural language models learn word representations, or embeddings, that capture rich linguistic and conceptual information. Here we investigate the embeddings learned by \emph{neural machine translation models}, a recently-developed class of neural language model. We show that embeddings from translation models outperform those learned by monolingual models at tasks that require knowledge of both conceptual similarity and lexical-syntactic role. We further show that these effects hold when translating from both English to French and English to German, and argue that the desirable properties of translation embeddings should emerge largely independently of the source and target languages. Finally, we apply a new method for training neural translation models with very large vocabularies, and show that this vocabulary expansion algorithm results in minimal degradation of embedding quality. Our embedding spaces can be queried in an online demo and downloaded from our web page. Overall, our analyses indicate that translation-based embeddings should be used in applications that require concepts to be organised according to similarity and/or lexical function, while monolingual embeddings are better suited to modelling (nonspecific) inter-word relatedness.
\end{abstract}

\section{Introduction}

It is well known that word representations can be learned from the distributional patterns in corpora. Originally, such representations were constructed by counting word co-occurrences, so that the features in one word's representation corresponded to other words~\citep{landauer1997solution,turney2010frequency}. Neural language models, an alternative method for learning word representations, use language data to optimise (latent) features with respect to a language modelling objective. The objective can be to predict either the next word given the initial words of a sentence~\citep{Bengio2003lm,mnih2009scalable,collobert2008unified}, or simply a nearby word given a single cue word~\citep{mikolov2013distributed,Pennington2014}. The representations learned by neural models (sometimes called \emph{embeddings}) perform very effectively when applied as pre-trained features in a range of NLP applications and tasks~\citep{baroni2014don}. 

Despite these clear results, it is not well understood how the architecture of neural models affects the information encoded in their embeddings. Here we contribute to this understanding by considering the embeddings learned by architectures with a very different objective function: \emph{neural machine translation} (NMT) \emph{models}. NMT models have recently emerged as an alternative to statistical, phrase-based translation models, and are beginning to achieve impressive translation performance~\citep{kalchbrenner13emnlp,devlin2014fast,Sutskever2014sequence}.

We show that NMT models are not only a potential new direction for machine translation, but are also an effective means of learning word embeddings. Specifically, translation-based embeddings encode information relating to conceptual similarity (rather than non-specific relatedness or association) and lexical syntactic role more effectively than embeddings from monolingual neural language models. We demonstrate that these properties persist when translating between different language pairs (English-French and English-German). Further, based on the observation of subtle language-specific effects in the embedding spaces, we conjecture as to why similarity dominates over other semantic relations in translation embedding spaces. Finally, we discuss a potential limitation of the application of NMT models for embedding learning - the computational cost of training large vocabularies of embeddings - and show that a novel method for overcoming this issue preserves the aforementioned properties of translation-based embeddings. 

% We show that the precise information encoFinally, we consider ways to combine the distinct information encoded in language-model embeddings and those from translation embeddings. And we consider ways to overcome two important limitations of neural translation models as embedding-learning architectures: the computational difficulty in learning large vocabularies of embeddings, and the requirement for sentence-aligned bilingual training corpora.   

\section{Learning Embeddings with Neural Language Models}

All neural language models, including NMT models, learn real-valued embeddings (of specified dimension) for words in some pre-specified vocabulary, \(V\), covering many or all words in their training corpus. At each training step, a `score' for the current training example (or batch) is computed based on the embeddings in their current state. This score is compared to the model's objective function, and the error is backpropagated to update both the model weights (affecting how the score is computed from the embeddings) and the embedding features themselves. At the end of this process, the embeddings should encode information that enables the model to optimally satisfy its objective.     

\subsection{Monolingual Models}

In the original neural language model~\citep{Bengio2003lm} and subsequent
variants \citep{collobert2008unified}, training examples consist of an ordered sequence of $n$ words,
with the model trained to predict the $n$-th word given the first $n-1$
words. The model first represents the input as an ordered sequence of embeddings,
which it transforms into a single fixed length `hidden' representation, generally by concatenation and non-linear projection.
Based on this representation, a probability distribution is
computed over the vocabulary, from which the model can sample a guess for the
next word. The model weights and embeddings are updated to maximise the
probability of correct guesses for all sentences in the
training corpus. 
 
More recent work has shown that high quality word embeddings can be learned via simpler models
with no nonlinear hidden layer ~\citep{mikolov2013distributed,Pennington2014}. Given a single word or unordered window of words in the corpus, these models predict which words will occur nearby. For each word \( w\) in \(V\), a list of training cases \({(w,c) : c \in V }\) is extracted from the training corpus according to some algorithm.  For instance, in the \emph{skipgram}
approach ~\citep{mikolov2013distributed}, for each `cue word' \(w\) the `context words' \(c\) are sampled
from windows either side of tokens of \(w\) in the corpus (with \(c\) more
likely to be sampled if it occurs closer to \(w\)).\footnote{
    Subsequent variants use different algorithms for selecting the
    \((w,c)\) from the training corpus \citep{Hill2014EMNLP,levy2014dependency}
} For each \(w\) in \( V\), the model initialises both a
cue-embedding, representing the \(w\) when it occurs as a cue-word, and a
context-embedding, used when \(w\) occurs as a context-word. For a cue word
\(w\), the model uses the corresponding cue-embedding and all
context-embeddings to compute a probability distribution over \(V\) that
reflects the probability of a word occurring in the context of \(w\). When a
training example \((w,c)\) is observed, the model updates both the cue-word
embedding of \(w\) and the context-word embeddings in order to increase the
conditional probability of \(c\). 

\subsection{Bilingual Representation-learning Models}
Various studies have demonstrated that word representations can also be effectively learned from bilingual corpora, aligned at the document, paragraph or word level~\citep{haghighi2008learning,vulic2011identifying,mikolov2013exploiting,Hermann:2014:ICLR,Chandar}. These approaches aim to represent the words from two (or more) languages in a common vector space so that words in one language are close to words with similar or related meanings in the other. The resulting multilingual embedding spaces have been effectively applied to bilingual lexicon extraction~\citep{haghighi2008learning,vulic2011identifying,mikolov2013exploiting} and document classification~\citep{klementiev2012inducing,Hermann:2014:ICLR,Chandar,Kocisky:2014}.

We focus our analysis on two representatives of this class of (non-NMT) bilingual model. The first is that of \cite{Hermann:2014:ICLR}, whose embeddings improve on the performance of~\cite{Klementiev} in document classification applications. As with the NMT models introduced in the next section, this model can be trained directly on bitexts aligned only at the sentence rather than word level. When training, for aligned sentences \(S_E\) and \(S_F\) in different languages, the model computes representations \(R_E\) and \(R_F\) by summing the embeddings of the words in \(S_E\) and \(S_F\) respectively. The embeddings are then updated to minimise the divergence between \(R_E\) and \(R_F\) (since they convey a common meaning). A noise-contrastive loss function ensures that the model does not arrive at trivial (e.g. all zero) solutions to this objective.~\cite{Hermann:2014:ICLR} show that, despite the lack of prespecified word alignments, words in the two languages with similar meanings converge in the bilingual embedding space.\footnote{The models of \cite{Chandar} and \cite{Hermann:2014:ICLR} both aim to minimise the divergence between source and target language sentences represented as sums of word embeddings. Because of these similarities, we do not compare with both in this paper.}

The second model we examine is that of \cite{faruqui2014improving}. Unlike the models described above, \cite{faruqui2014improving} showed explicitly that projecting word embeddings from two languages (learned independently) into a common vector space can favourably influence the orientation of word embeddings when considered in their monolingual subspace; i.e relative to other words in their own language.  In contrast to the other models considered in this paper, the approach of \cite{faruqui2014improving} requires bilingual data to be aligned at the word level.

\subsection{Neural Machine Translation Models}

The objective of NMT is to generate an appropriate sentence in a target
language \(S_t\)  given a sentence \(S_s\) in the source language~\citep[see,
e.g.,][]{kalchbrenner13emnlp,Sutskever2014sequence}. As a by-product of learning to meet this objective, NMT models learn distinct sets of embeddings for the vocabularies \(V_ s\) and \(V_t\) in the source and target languages respectively.

Observing a training case \((S_s, S_t)\), these models represent \(S_s\) as an ordered sequence of embeddings of words from \(V_s\). The sequence for \(S_s\) is then encoded into a single representation \(R_S\).\footnote{Alternatively, subsequences (phrases) of \(S_s\) may be encoded at this stage in place of the whole sentence~\citep{Bahdanau2014}.} Finally, by referencing the embeddings in \(V_t\), \(R_S\) and a representation of what has been generated thus far, the model decodes a sentence in the target language word by word. If at any stage the decoded word does not match the corresponding word in the training target \(S_t\), the error is recorded. The weights and embeddings in the model, which together parameterise the encoding and decoding process, are updated based on the accumulated error once the sentence decoding is complete. 

Although NMT models can differ in their low-level architecture~\citep{kalchbrenner13emnlp,Cho2014,Bahdanau2014}, the translation objective exerts similar pressure on the embeddings in all cases. The source language embeddings must be such that the model can combine them to form single representations for ordered sequences of multiple words (which in turn must enable the decoding process). The target language embeddings must facilitate the process of decoding these representations into correct target-language sentences.    

\section{Experiments}

To learn translation-based embeddings, we trained two different NMT models. The first is the RNN encoder-decoder~\citep[\emph{RNNenc},][]{Cho2014}, which uses a recurrent-neural-network to encode all of the source sentence into a single vector on which the decoding process is conditioned. The second is the \emph{RNN Search} architecture~\citep{Bahdanau2014}, which was designed to overcome limitations exhibited by the RNN encoder-decoder when translating very long sentences. RNN Search includes a \emph{attention} mechanism, an additional feed-forward network that learns to attend to different parts of the source sentence when decoding each word in the target sentence.\footnote{Access to source code and limited GPU time prevent us from training and evaluating the embeddings from other NMT models such as that of~\citep{kalchbrenner13emnlp},~\citep{devlin2014fast} and \cite{Sutskever2014sequence}. The underlying principles of encoding-decoding also apply to these models, and we expect the embeddings would exhibit similar properties to those analysed here.} Both models were trained on a 348m word corpus of English-French sentence pairs or a 91m word corpus of English-German sentence pairs.\footnote{These corpora were produced from the WMT ’14 parallel data after conducting the data-selection procedure described by~\cite{Cho2014}. } 

To explore the properties of bilingual embeddings learned via objectives other than direct translation, we trained the \emph{BiCVM} model of~\cite{Hermann:2014:ICLR} on the same data, and also downloaded the projected embeddings of~\cite{faruqui2014improving}, \emph{FD}, trained on a bilingual corpus of comparable size (\(\approx 300\) million words per language).\footnote{Available from \url{http://www.cs.cmu.edu/~mfaruqui/soft.html}. The available embeddings were trained on English-German aligned data, but the authors report similar to for English-French.} Finally, for an initial comparison with monolingual models, we trained a conventional skipgram model~\citep{mikolov2013distributed} and its \emph{Glove} variant~\citep{Pennington2014} for the same number of epochs on the English half of the bilingual corpus. 

To analyse the effect on embedding quality of increasing the quantity of training data, we then trained the monolingual models on increasingly large random subsamples of Wikipedia text (up to a total of 1.1bn words). Lastly, we extracted embeddings from a full-sentence language model~\citep[\emph{CW},][]{collobert2008unified}, which was trained for several months on the same Wikipedia 1bn word corpus. Note that increasing the volume of training data for the bilingual (and NMT) models was not possible because of the limited size of available sentence-aligned bitexts. 
 
\subsection{Similarity and relatedness modelling}

As in previous studies~\citep{Agirre2009,Bruni2014,baroni2014don}, our initial evaluations involved calculating pairwise (cosine) distances between embeddings and correlating these distances with (gold-standard) human judgements of the strength of relationships between concepts. For this we used three different gold standards: WordSim-353~\citep{Agirre2009}, MEN~\citep{Bruni2014} and SimLex-999~\citep{hill2014simlex}. Importantly, there is a clear distinction between WordSim-353 and MEN, on the one hand, and SimLex-999, on the other, in terms of the semantic relationship that they quantify. For both WordSim-353 and MEN, annotators were asked to rate how \emph{related} or \emph{associated} two concepts are. Consequently, pairs such as [\emph{clothes}-\emph{closet}], which are clearly related but ontologically dissimilar, have high ratings in WordSim-353 and MEN. In contrast, such pairs receive a low rating in SimLex-999, where only genuinely \emph{similar} concepts, such as [\emph{coast}- \emph{shore}], receive high ratings. 

To reproduce the scores in SimLex-999, models must thus distinguish pairs that are similar from those that are merely related. In particular, this requires models to develop sensitivity to the distinction between synonyms (similar) and antonyms (often strongly related, but highly dissimilar).\footnote{For a more detailed discussion of the similarity/relatedness distinction, see~\citep{hill2014simlex}.}

Table~\ref{table:perf} shows the correlations of NMT (English-French) embeddings, other bilingually-trained embeddings and monolingual embeddings with these three lexical gold-standards. NMT outperform monolingual embeddings, and, to a lesser extent, the other bilingually trained embeddings, on SimLex-999. However, this clear advantage is not observed on MEN and WordSim-353, where the projected embeddings of~\cite{faruqui2014improving}, which were tuned for high performance on WordSim-353, perform best. Given the aforementioned differences between the evaluations, this suggests that bilingually-trained embeddings, and NMT based embeddings in particular, better capture similarity, whereas monolingual embedding spaces are orientated more towards relatedness. 

\begin{table}[t]
\begin{center}
\begin{tabular}{r c | r  r  r  | r r | r r |}
\multicolumn{2}{c|}{~} &\multicolumn{3}{c|}{Monolingual models}  & \multicolumn{2}{c|}{\small Biling. models} & \multicolumn{2}{c|}{NMT models}\\ 
    \multicolumn{2}{c|}{~} &\bf Skipgram &\bf Glove &\bf CW & \bf FD & \bf BiCVM & \bf RNNenc &\bf RNNsearch \\ 
\hline
WordSim-353   & \(\rho\) & 0.52 & 0.55 & 0.51 &{\bf  0.69} & 0.50 &   0.57 &  0.58 \\
MEN & \(\rho\) & 0.44 & 0.71 & 0.60 & {\bf 0.78} & 0.45 &  0.63 &  0.62  \\
\hdashline
SimLex-999 & \(\rho\) & 0.29 & 0.32 & 0.28 & 0.39 & 0.36 &   {\bf 0.52} &  0.49 \\
SimLex-333 & \(\rho\) &  018&0.18  &0.07  & 0.24  &  0.34 & { \bf 0.49}   & 0.45   \\
TOEFL & \(\%\) & 0.75 & 0.78 & 0.64 & 0.84 & 0.87 &  {\bf 0.93} &  {\bf 0.93} \\
Syn/antonym & \(\%\) & 0.69  & 0.72  &  0.75 & 0.76 & 0.70 &  {\bf 0.79} & 0.74 \\
\end{tabular}
\caption{ NMT embeddings (RNNenc and RNNsearch) clearly outperform alternative embedding-learning architectures on tasks that require modelling similarity (below the dashed line), but not on tasks that reflect relatedness. Bilingual embedding spaces learned without the translation objective are somewhere between these two extremes.}
\label{table:perf}
\end{center}
\vspace{-5mm}
\end{table}

\begin{table}[t]
\begin{center}
\begin{tabular}{r | r  r  r | r r | r r}
&\bf Skipgram &\bf Glove &\bf CW&\bf FD &\bf BiCVM  &\bf RNNenc &\bf RNNsearch \\ 
\hline
\emph{teacher}  & {\small \emph{vocational}} &  {\small \emph{student}} 
& {\small \emph{student}} &{\small \emph{elementary}} & {\small  \emph{faculty}} & {\small \emph{professor}}  & {\small \emph{instructor}} \\ 
 & {\small \emph{in-service}} &  {\small \emph{pupil}} 
& {\small \emph{tutor}} & {\small \emph{school}}& {\small  \emph{professors}} & {\small \emph{instructor}}  & {\small \emph{professor}} \\ 
 & {\small \emph{college}} &  {\small \emph{university}} 
& {\small \emph{mentor}} & {\small \emph{classroom}}& {\small \emph{teach}}& {\small \emph{trainer}}  & {\small \emph{educator}} \\ 
\hdashline
\emph{eaten}  & {\small \emph{spoiled}} &  {\small \emph{cooked}} 
&  {\small \emph{baked}} &{\small \emph{ate}}& {\small  \emph{eating}}& {\small \emph{ate}} & {\small \emph{ate}} \\ 
  & {\small \emph{squeezed}} &  {\small \emph{eat}} 
&  {\small \emph{peeled}} &{\small \emph{meal}}& {\small \emph{eat}}& {\small \emph{consumed}} & {\small \emph{consumed}} \\ 
  & {\small \emph{cooked}} &  {\small \emph{eating}} 
&  {\small \emph{cooked}} &{\small \emph{salads}}& {\small \emph{baking}}& {\small \emph{tasted}} & {\small \emph{eat}} \\ 
\hdashline
\emph{Britain}  & {\small \emph{Northern}} &  {\small \emph{Ireland}} 
& {\small \emph{Luxembourg}} &{\small \emph{UK}}& {\small \emph{UK}} & {\small  \emph{UK}} & {\small \emph{England}} \\ 
& {\small \emph{Great}} &  {\small \emph{Kingdom}} 
& {\small \emph{Belgium}} &{\small \emph{British}}& {\small \emph{British}} & {\small  \emph{British}} & {\small \emph{UK}} \\ 
 & {\small \emph{Ireland}} &  {\small \emph{Great}} 
& {\small \emph{Madrid}} &{\small \emph{London}}& {\small \emph{England}} & {\small  \emph{America}} & {\small \emph{Syria}} \\

\end{tabular}
\caption{Nearest neighbours (excluding plurals) in the embedding spaces of different models. All models were trained for 6 epochs on the translation corpus except CW and FD (as noted previously). NMT embedding spaces are oriented according to similarity, whereas embeddings learned by monolingual models are organized according to relatedness. The other bilingual model BiCVM also exhibits a notable focus on similarity.}
\label{table:neigh}
\end{center}
\vspace{-5mm}
\end{table}

To test this hypothesis further, we ran three more evaluations designed to probe the sensitivity of models to similarity as distinct from relatedness or association. In the first, we measured performance on SimLex-Assoc-333 ~\citep{hill2014simlex}. This evaluation comprises the 333 most related pairs in SimLex-999, according to an independent empirical measure of relatedness (free associate generation~\citep{nelson2004university}). Importantly, the pairs in SimLex-Assoc-333, while all strongly related, still span the full range of similarity scores.\footnote{The most dissimilar pair in SimLex-Assoc-333 is [\emph{shrink,grow}] with a score of 0.23. The highest is [\emph{vanish},\emph{disappear}] with 9.80.} Therefore, the extent to which embeddings can model this data reflects their sensitivity to the similarity (or dissimilarity) of two concepts, even in the face of a strong signal in the training data that those concepts are related.    

The TOEFL synonym test is another similarity-focused evaluation of embedding spaces. This test contains 80 cue words, each with four possible answers, of which one is a correct synonym~\citep{landauer1997solution}. We computed the proportion of questions answered correctly by each model, where a model's answer was the nearest (cosine) neighbour to the cue word in its vocabulary.\footnote{To control for different vocabularies, we restricted the effective vocabulary of each model to the intersection of all model vocabularies, and excluded all questions that contained an answer outside of this intersection.} Note that, since TOEFL is a test of synonym recognition, it necessarily requires models to recognise similarity as opposed to relatedness.  

Finally, we tested how well different embeddings enabled a supervised classifier to distinguish between synonyms and antonyms, since synonyms are necessarily similar and people often find antonyms, which are necessarily dissimilar, to be strongly associated. For 744 word pairs hand-selected as either synonyms or antonyms,\footnote{Available online at \url{http://www.cl.cam.ac.uk/~fh295/}.} we presented a Gaussian SVM with the concatenation of the two word embeddings. We evaluated accuracy using 10-fold cross-validation. 

As shown in Table~\ref{table:perf}, with these three additional similarity-focused tasks we again see the same pattern of results. NMT embeddings outperform other bilingually-trained embeddings which in turn outperform monolingual models. The difference is particularly striking on SimLex-Assoc-333, which suggests that the ability to discern similarity from relatedness (when relatedness is high) is perhaps the most clear distinction between the bilingual spaces and those of monolingual models. 

These conclusions are also supported by qualitative analysis of the various embedding spaces. As shown in Table~\ref{table:neigh}, in the NMT embedding spaces the nearest neighbours (by cosine distance) to concepts such as \emph{teacher} are genuine synonyms such as \emph{professor} or \emph{instructor}. The bilingual objective also seems to orientate the non-NMT embeddings towards semantic similarity, although some purely related neighbours are also oberved. In contrast, in the monolingual embedding spaces the neighbours of \emph{teacher} include  highly related but dissimilar concepts such as \emph{student} or \emph{college}. 

%\footnote{Readers can inspect nearest neighbours in each embedding space using our web demo.} 
 
\subsection{Importance of training data quantity}

\begin{figure*}[h]
\includegraphics[width = \textwidth,clip=True,trim=0 10 0 10]{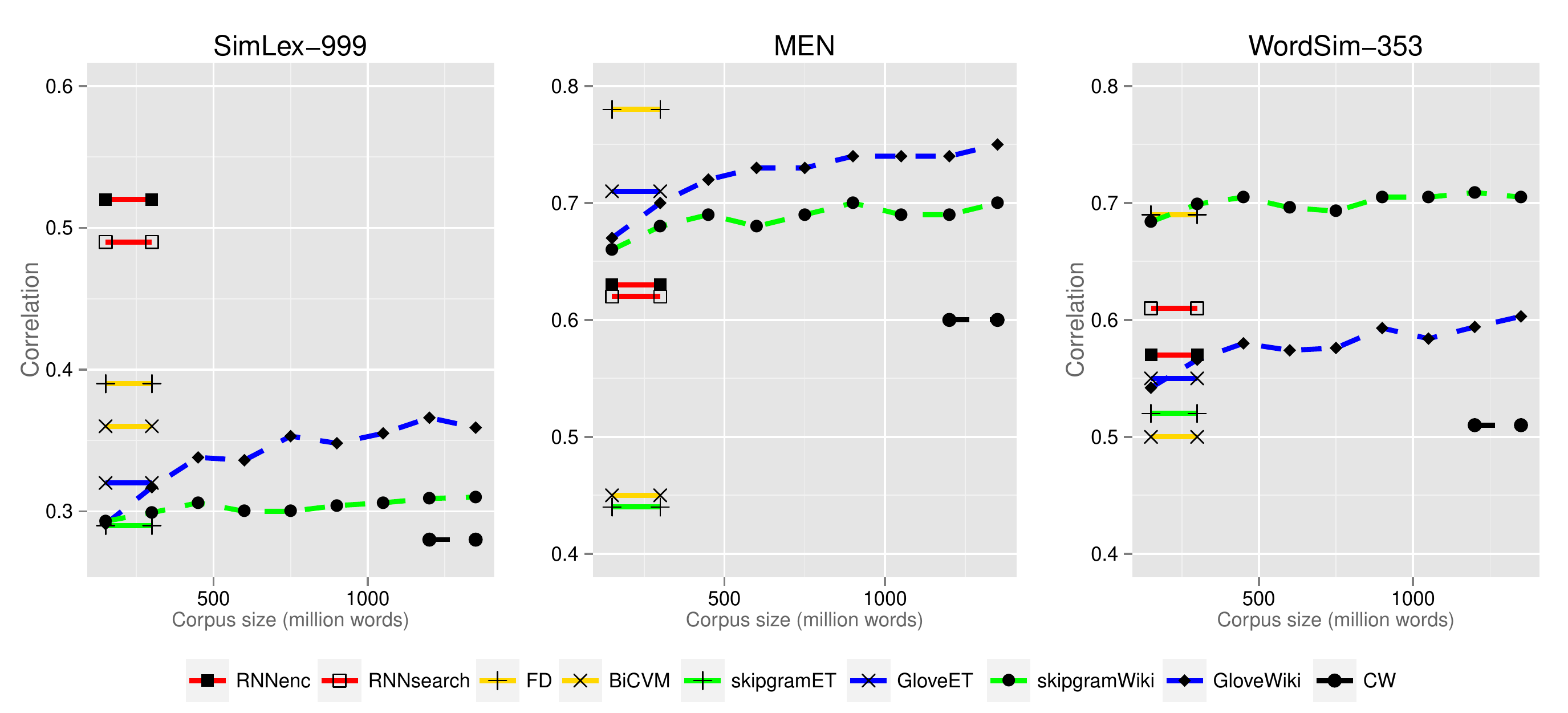}
\vspace{-4mm}
\caption{The effect of increasing the amount of training data on the quality of monolingual embeddings, based on similarity-based evaluations (SimLex-999) and two relatedness-based evaluations (MEN and WordSim-353). \emph{ET} in the legend indicates models trained on the English half of the translation corpus. \emph{Wiki} indicates models trained on Wikipedia.}
\label{fig:size}
\end{figure*}

In previous work, monolingual models were trained on corpora many times larger than the English half of our parallel translation corpus. Indeed, the ability to scale to large quantities of training data was one of the principal motivations behind the skipgram architecture~\citep{mikolov2013distributed}. To check if monolingual models simply need more training data to capture similarity as effectively as bilingual models, we therefore trained them on increasingly large subsets of Wikipedia.\footnote{We did not do the same for our translation models because sentence-aligned bilingual corpora of comparable size do not exist.} As shown in Figure~\ref{fig:size}, this is not in fact the case. The performance of monolingual embeddings on similarity tasks remains well below the level of the NMT embeddings and somewhat lower than the non-MT bilingual embeddings as the amount of training data increases. 

\subsection{Analogy Resolution}

Lexical analogy questions have been used as an alternative way of evaluating word representations. In this task, models must identify the correct answer (\emph{girl}) when presented with analogy questions such as `\emph{man} is to \emph{boy} as \emph{woman} is to ?'. It has been shown that Skipgram-style models are surprisingly effective at answering such questions~\citep{mikolov2013distributed}. This is because, if \( \bf m, b \) and \( \bf w\) are skigram-style embeddings for \emph{man}, \emph{boy} and \emph{woman} respectively, the correct answer is often the nearest neighbour in the vocabulary (by cosine distance) to the vector \( \bf v = w + b - m \). 

We evaluated embeddings on analogy questions using the same vector-algebra method as~\citet{mikolov2013distributed}. As in the previous section, for fair comparison we excluded questions containing a word outside the intersection of all model vocabularies, and restricted all answer searches to this reduced vocabulary. This left 11,166 analogies. Of these, 7219 are classed as `syntactic', in that they exemplify mappings between parts-of-speech or syntactic roles (e.g. \emph{fast} is to \emph{fastest} as \emph{heavy} is to \emph{heaviest}), and 3947 are classed as `semantic` (\emph{Ottawa} is to \emph{Canada} as \emph{Paris} is to \emph{France}), since successful answering seems to rely on some (world) knowledge of the concepts themselves. 

\begin{figure*}[ht]

\includegraphics[width = \textwidth,clip=True,trim=0 10 0 10]{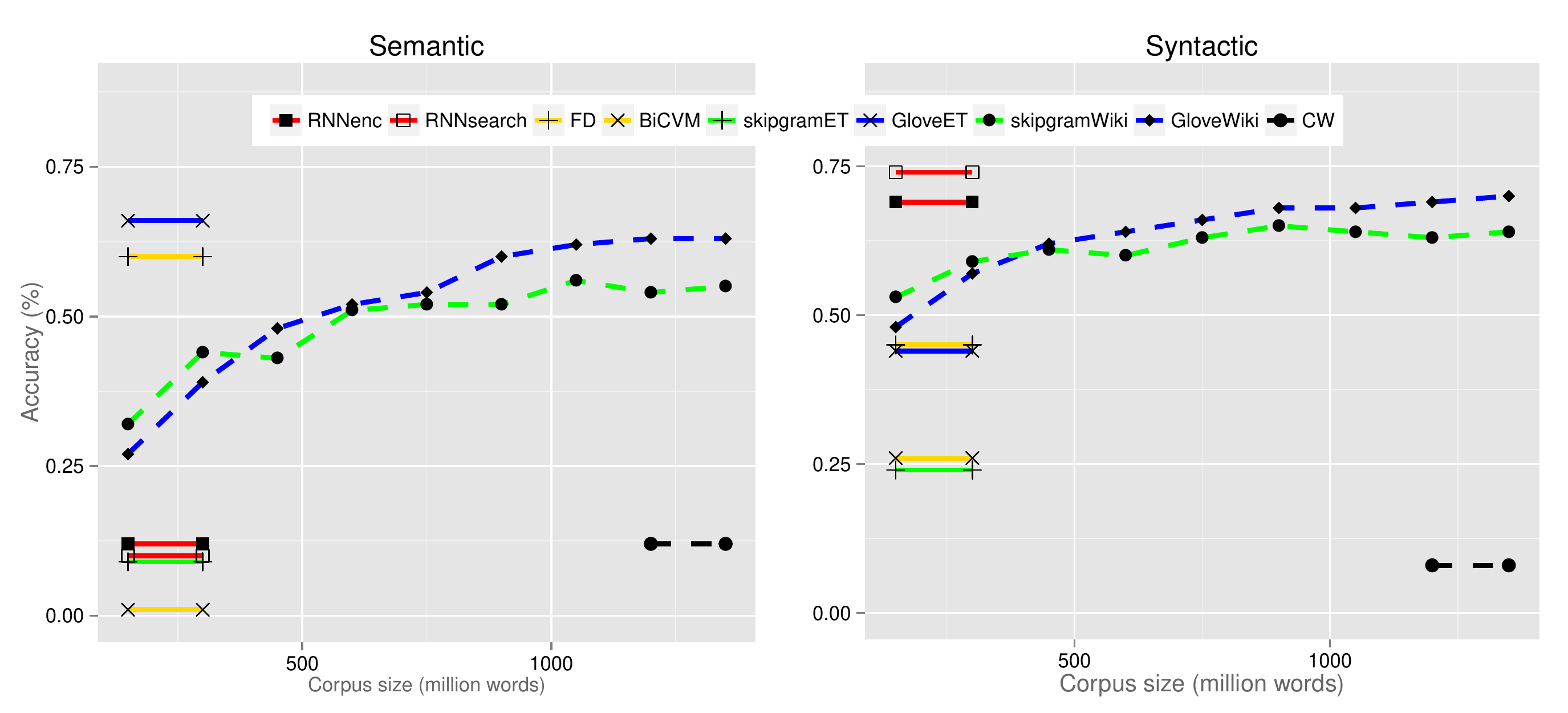}

\vspace{-4mm}
\caption{Translation-based embeddings perform best on syntactic analogies (\emph{run,ran: hide, hid}). Monolingual skipgram/Glove models are better at semantic analogies (\emph{father, man; mother, woman})}
\label{fig:analogy}

\end{figure*}

As shown in Fig.~\ref{fig:analogy}, NMT embeddings yield relatively poor answers to semantic analogy questions compared with monolingual embeddings and the bilingual embeddings \emph{FD} (which are projections of similar monolingual embeddings).\footnote{The performance of the FD embeddings on this task is higher than that reported by~\cite{faruqui2014improving} because we search for answers over a smaller total candidate vocabulary.} It appears that the translation objective prevents the embedding space from developing the same linear, geometric regularities as skipgram-style models with respect to semantic organisation. This also seems to be true of the embeddings from the full-sentence language model \emph{CW}. Further, in the case of the Glove and FD models this advantage seems to be independent of both the domain and size of the training data, since embeddings from these models trained on only the English half of the translation corpus still outperform the translation embeddings. 

On the other hand, NMT embeddings are effective for answering syntactic analogies using the vector algebra method. They perform comparably to or even better than monolingual embeddings when trained on less data (albeit bilingual data). It is perhaps unsurprising that the translation objective incentivises the encoding of a high degree of lexical syntactic information, since coherent target-language sentences could not be generated without knowledge of the parts-of-speech, tense or case of its vocabulary items. The connection between the translation objective and the embedding of lexical syntactic information is further supported by the fact that embeddings learned by the bilingual model BiCVM do not perform comparably on the syntactic analogy task. In this model, sentential semantics is transferred via a bag-of-words representation, presumably rendering the precise syntactic information less important.

When considering the two properties of NMT embeddings highlighted by these experiments, namely the encoding of semantic similarity and lexical syntax, it is worth noting that items in the similarity-focused evaluations of the previous section (SimLex-999 and TOEFL) consist of word groups or pairs that have identical syntactic role. Thus, even though lexical semantic information is in general pertinent to conceptual similarity~\citep{levy2014dependency}, the lexical syntactic and conceptual properties of translation embeddings are in some sense independent of one another.

\section{Effect of Target Language}

To better understand why a translation objective yields embedding spaces with particular properties, we trained the RNN Search architecture to translate from English to German. 

\begin{table}[ht]
\begin{center}
\begin{tabular}{r c | m{0.9cm}  m{0.9cm}  r c c c }
    \multicolumn{2}{c|}{~} &\bf  \small EN-FR &\bf  \small EN-DE &  &  `earned' & `castle' & `money'\\ 
\cline{1-4} \cline{6-8}
WordSim-353   & \(\rho\) & 0.60 & \bf 0.61 &   & {\small \emph{gained}} & {\small \emph{chateau}} & {\small \bf \emph{ silver}} \\
MEN & \(\rho\) & 0.61 & \bf 0.62 \bf & \bf \small  EN-FR  & {\small \bf \emph{won}} & {\small \emph{palace}} & {\small \emph{funds}} \\
SimLex-999 & \(\rho\) & 0.49 &  \bf 0.50 &  & {\small \emph{acquired}}  & {\small \emph{fortress}}  & {\small \emph{cash}} \\
\cline{6-8}
SimLex-Assoc-333 & \(\rho\) & 0.45  & \bf 0.47   &   &  &   \\
TOEFL & \(\%\) & 0.90 & \bf 0.93  & &  {\small \emph{gained}}&  {\small \emph{chateau}} &  {\small \emph{funds}} \\ 
Syn/antonym & \(\%\) & \bf 0.72 &  0.70  &  \bf \small EN-DE   &  {\small \emph{deserved}}    &  {\small \emph{palace}}    &  {\small \emph{cash}} \\ 
Syntactic analogies & \(\%\) & \bf 0.73 & 0.62   &  &  {\small \emph{accummulated}}   &  {\small \bf \emph{ padlock}}   &  {\small \emph{resources}} \\  
Semantic analogies & \(\%\) & 0.10 & \bf  0.11  \\
\end{tabular}
\caption{Comparison of embeddings learned by RNN Search models translating between English-French (EN-FR) and English-German (EN-DE) on all semantic evaluations (left) and nearest neighbours of selected cue words (right). Bold italics indicate target-language-specific effects. Evaluation items and vocabulary searches were restricted to words common to both models. }
\label{table:de}
\end{center}
\vspace{-5mm}
\end{table}

As shown in Table~\ref{table:de} (left side), the performance of the source (English) embeddings learned by this model was comparable to that of those learned by the English-to-French model on all evaluations, even though the English-German training corpus (91 million words) was notably smaller than the English-French corpus (348m words). This evidence shows that the desirable properties of translation embeddings highlighted thus far are not particular to English-French translation, and can also emerge when translating to a different language family, with different word ordering conventions.     

\section{Overcoming the Vocabulary Size Problem}

A potential drawback to using NMT models for learning word embeddings is the computational cost of training such a model on large vocabularies. To generate a target language sentence, NMT models repeatedly compute a softmax distribution over the target vocabulary. This computation scales with vocabulary size and must be repeated for each word in the output sentence, so that training models with large output vocabularies is challenging. Moreover, while the same computational bottleneck does not apply to the encoding process or source vocabulary, there is no way in which a translation model could learn a high quality source embedding for a word if the plausible translations were outside its vocabulary. Thus, limitations on the size of the target vocabulary effectively limit the scope of NMT models as representation-learning tools. This contrasts with the shallower monolingual and bilingual representation-learning models considered in this paper,  which efficiently compute a distribution over a large target vocabulary using either a hierarchical softmax~\citep{morin2005hierarchical} or approximate methods such as negative sampling~\citep{mikolov2013distributed,Hermann:2014:ICLR}, and thus can learn large vocabularies of both source and target embeddings.

A recently proposed solution to this problem enables NMT models to be trained with larger target vocabularies (and hence larger meaningful source vocabularies) at comparable computational cost to training with a small target vocabulary~\citep{Jean}. The algorithm uses (biased) importance sampling~\citep{Bengio+Senecal-2003-small} to approximate the probability distribution of words over a large target vocabulary with a finite set of distributions over subsets of that vocabulary. Despite this element of approximation in the decoder, extending the effective target vocabulary in this way significantly improves translation performance, since the model can make sense of more sentences in the training data and encounters fewer unknown words at test time. In terms of representation learning, the method provides a means to scale up the NMT approach to vocabularies as large as those learned by monolingual models. However, given that the method replaces an exact calculation with an approximate one, we tested how the quality of source embeddings is affected by scaling up the target language vocabulary in this way. 

\begin{table}[t]
\begin{center}
\begin{tabular}{r c | c c c c }
    \multicolumn{2}{c|}{~} &\bf RNN Search &\bf RNN Search & \bf RNN Search-LV  & \bf RNN Search-LV  \\ 
 \multicolumn{2}{c|}{~} &\bf \small EN-FR &\bf \small  EN-DE & \bf  \small EN-FR & \bf \small EN-DE \\ 
\hline
WordSim-353   & \(\rho\) & 0.60 & \bf 0.61 & 0.59 & 0.57  \\
MEN & \(\rho\) & 0.61 & \bf 0.62 & \bf 0.62 & 0.61 \\
SimLex-999 & \(\rho\) & 0.49 & 0.50 & \bf  0.51 & 0.50  \\
SimLex-Assoc-333 & \(\rho\) & 0.45  & \bf 0.47  & \bf 0.47  & 0.46   \\
TOEFL & \(\%\) & 0.90 & 0.93 & 0.93 & \bf 0.98  \\
Syn/antonym & \(\%\) & 0.72 &  0.70 & \bf 0.74 & 0.71 \\
Syntactic analogies & \(\%\) & \bf 0.73 &  0.62 & 0.71 & 0.62\\
Semantic analogies & \(\%\) & 0.10 &  0.11 & 0.08 & \bf 0.13\
\end{tabular}
\caption{Comparison of embeddings learned by the original (RNN Search - 30k French words, 50k German words) and extended-vocabulary (RNN Search-LV -500k words) models translating from English to French (EN-FR) and from English to German (EN-DE). For fair comparisons, all evaluations were restricted to the intersection of all model vocabularies.}
\label{table:ex}
\end{center}
\vspace{-5mm}
\end{table}

As shown in Table~\ref{table:ex}, there is no significant degradation of embedding quality when scaling to large vocabularies with using the approximate decoder. Note that for a fair comparison we filtered these evaluations to only include items that are present in the smaller vocabulary. Thus, the numbers do not directly reflect the quality of the additional 470k embeddings learned by the extended vocabulary models, which one would expect to be lower since they are words of lower frequency. All embeddings can be downloaded from \url{http://www.cl.cam.ac.uk/~fh295/}, and the embeddings from the smaller vocabulary models can be interrogated at \url{http://lisa.iro.umontreal.ca/mt-demo/embs/}.\footnote{A different solution to the rare-word problem was proposed by~\citep{luong2014addressing}. We do not evaluate the effects on the resulting embeddings of this method because we lack access to the source code.}

\section{How Similarity Emerges}
\label{section:exp}

Although NMT models appear to encode both conceptual similarity and syntactic information for any source and target languages, it is not the case that embedding spaces will always be identical. Interrogating the nearest neighbours of the source embedding spaces of the English-French and English-German models reveals occasional language-specific effects. As shown in Table~\ref{table:de} (right side), the neighbours for the word \emph{earned} in the English-German model are as one might expect, whereas the neighbours from the English-French model contain the somewhat unlikely candidate \emph{won}. In a similar vein, while the neighbours of the word \emph{castle} from the English-French model are unarguably similar, the neighbours from the English-German model contain the word \emph{padlock}.
 
These infrequent but striking differences between the English-German and English-French source embedding spaces indicate how similarity might emerge effectively in NMT models. Tokens of the French verb \emph{gagner} have (at least) two possible English translations (\emph{win} and \emph{earn}). Since the translation model, which has limited encoding capacity, is trained to map tokens of \emph{win} and \emph{earn} to the same place in the target embedding space, it is efficient to move these concepts closer in the source space. Since \emph{win} and \emph{earn} map directly to two different verbs in German, this effect is not observed. On the other hand, the English nouns \emph{castle} and \emph{padlock} translate to a single noun (\emph{Schloss}) in German, but different nouns in French. Thus, \emph{padlock} and \emph{castle} are only close in the source embeddings from the English-German model. 

Based on these considerations, we can conjecture that the following condition on the semantic configuration between two language is crucial to the effective induction of lexical similarity. 

\MyQuote{\emph{For \(s_1\) and \(s_2\) in the source language, there is some \(t\) in the target language such that there are sentences in the training data in which \(s_1\) translates to \(t\) and sentences in which \(s_2\) translates to \(t\).}}

{\centering \emph{if and only if} \\}

\MyQuote{\emph{\(s_1\) and \(s_2\) are semantically similar.}}

Of course, this condition is not true in general. However, we propose that the extent to which it holds over all possible word pairs corresponds to the quality of similarity induction in the translation embedding space. Note that strong polysemy in the target language, such as \emph{gagner = win, earn}, can lead to cases in which \(1\) is satisfied but \(2\) is not. The conjecture claims that these cases are detrimental to the quality of the embedding space (at least with regards to similarity). In practice, qualitative analyses of the embedding spaces and native speaker intuitions suggest that such cases are comparatively rare. Moreover, when such cases are observed, \(s_1\) and \(s_2\), while perhaps not similar, are not strongly dissimilar. This could explain why related but strongly dissimilar concepts such as antonym pairs do not converge in the translation embedding space. This is also consistent with qualitative evidence presented by ~\citep{faruqui2014improving} that projecting monolingual embeddings into a bilingual space orientates them to better reflect the synonymy/antonymy distinction.

\section{Conclusion}

In this work, we have shown that the embedding spaces from neural machine translation models are orientated more towards conceptual similarity than those of monolingual models, and that translation embedding spaces also reflect richer lexical syntactic information. To perform well on similarity evaluations such as SimLex-999, embeddings must distinguish information pertinent to what concepts \emph{are} (their function or ontology) from information reflecting other non-specific inter-concept relationships. Concepts that are strongly related but dissimilar, such as antonyms, are particularly challenging in this regard~\citep{hill2014simlex}. Consistent with the qualitative observation made by ~\cite{faruqui2014improving}, we suggested how the nature of the semantic correspondence between the words in languages enables NMT embeddings to distinguish synonyms and antonyms and, more generally, to encode the information needed to reflect human intuitions of similarity.   

The language-specific effects we observed in Section 4 suggest a potential avenue for improving translation and multi-lingual embeddings in future work. First, as the availability of fast GPUs for training grows, we would like to explore the embeddings learned by NMT models that translate between much more distant language pairs such as English-Chinese or English-Arabic. For these language pairs, the word alignment will less monotonic and may result in even more important semantic and syntactic information being encoded in the lexical representation. Further,  as observed by both~\cite{Hermann:2014:ICLR} and~\cite{faruqui2014improving}, the bilingual representation learning paradigm can be naturally extended to update representations based on correspondences between multiple languages (for instance by interleaving English-French and English-German training examples). Such an approach should smooth out language-specific effects, leaving embeddings that encode only language-agnostic conceptual semantics and are thus more generally applicable. Another related challenge is to develop smaller or less complex representation-learning tools that encode similarity with as much fidelity as NMT models but without the computational overhead. One promising approach for this is to learn word alignments and word embeddings jointly~\citep{Kocisky:2014}. This approach is effective for cross-lingual document classification, although the authors do evaluate the monolingual subspace induced by the model.\footnote{These embeddings are not publicly available and we were unable to re-train them using the source code.}

Not all word embeddings learned from text are born equal. Depending on the application, those learned by NMT models may have particularly desirable properties. For decades, distributional semantic models have aimed to exploit Firth's famous \emph{distributional hypothesis} to induce word meanings from (monolingual) text. However, the hypothesis also betrays the weakness of the monolingual distributional approach when it comes to learning humah-quality concept representations. For while it is undeniable that ``words which are similar in meaning appear in similar distributional contexts"~\citep{dist}, the converse assertion, which is what really matters, is only sometimes true.

\subsubsection*{Acknowledgments}
The authors would like to thank the developers of
Theano~\cite{bergstra+al:2010-scipy,Bastien-Theano-2012}.  We acknowledge the
support of the following agencies for research funding and computing support:
St John's College Cambridge, NSERC, Calcul Qu\'{e}bec, Compute Canada, the Canada Research Chairs and CIFAR.

\bibliography{iclr2015}
\bibliographystyle{iclr2015}

\end{document}